\documentclass[lettersize,journal]{IEEEtran}
\usepackage{amsmath,amsfonts}
\usepackage{algorithmic}
\usepackage{algorithm}
\usepackage{array}
\usepackage[caption=false,font=normalsize,labelfont=sf,textfont=sf]{subfig}
\usepackage{textcomp}
\usepackage{stfloats}
\usepackage{url}
\usepackage{verbatim}
\usepackage{graphicx}
\usepackage{cite}

\usepackage{soul}
\usepackage{tablefootnote}
\usepackage[normalem]{ulem}
\usepackage{multirow}
\usepackage{makecell}
\usepackage{soul}
\usepackage{enumitem}
\usepackage{adjustbox}
\usepackage{amsmath}
\usepackage{amssymb}
\usepackage{array}
\usepackage{graphicx}
\usepackage{comment}
\usepackage{booktabs}
\usepackage{xcolor,colortbl} 
\usepackage{algorithm}
\usepackage{algorithmic}
\DeclareMathOperator*{\argmax}{arg\,max}

\usepackage{bbm}
\usepackage{tabularx}
\usepackage{hyperref}
\newcommand{\real}[1]{{\mathbb{R}^{{#1}}}}
\usepackage{graphicx}
\usepackage{arydshln}
\usepackage{array}
\usepackage{booktabs}
\usepackage{pifont}

\begin{document}
% Mark sections of captions for referring to divisions of figures
\newcommand{\tocite}[1]{{[\hl{CITE: #1]}}}
\newcommand{\todo}[1]{{[\hl{TODO: #1}]}}
\newcommand{\yifan}[1]{{[\hl{Yifan: #1}]}}

\newcommand{\modelnameshort}{\textsc{Mudern}}
\newcommand{\modelname}{\textbf{Mu}lti-passage \textbf{D}iscourse-aware \textbf{E}ntailment \textbf{R}easoning \textbf{N}etwork}
\newcommand{\sota}{{state-of-the-art}}
\newcommand{\ethree}{$\text{E}^{3}$}
\newcommand{\emt}{EMT}
\newcommand{\discern}{\textsc{Discern}}
\newcommand{\mprbt}{\textsc{MP}-RoBERTa}
\newcommand{\Fbleu}{$\mathrm{F1}_\text{BLEU}$}
\newcommand{\Fbleuone}{$\mathrm{F1}_\text{BLEU1}$}
\newcommand{\Fbleufour}{$\mathrm{F1}_\text{BLEU4}$}
\newcommand{\micro}{Micro Acc.}
\newcommand{\macro}{Macro Acc.}
\newcommand{\orsharc}{OR-ShARC}
\newcommand{\sharc}{ShARC}

\title{Open-Retrieval Conversational Machine Reading}

\author{Yifan Gao, Jingjing Li, Chien-Sheng Wu, Michael R. Lyu~\IEEEmembership{Fellow,~IEEE}, and Irwin King~\IEEEmembership{Fellow,~IEEE}
        % <-this % stops a space
\thanks{Yifan Gao, Jingjing Li, Michael R. Lyu, and Irwin King are with Department of Computer Science and Engineering, The Chinese University of Hong Kong, Shatin, N.T., Hong Kong (e-mail: \{yfgao,lijj,lyu,king\}@cse.cuhk.edu.hk).}% <-this % stops a space
\thanks{Chien-Sheng Wu is with Salesforce Research, 415 Mission Street, San Francisco (e-mail: wu.jason@salesforce.com).}% <-this % stops a space
% \thanks{Manuscript received April 19, 2021; revised August 16, 2021.}% <-this % stops a space
% \thanks{preprint, under review.}
}

\markboth{Journal of \LaTeX\ Class Files,~Vol.~14, No.~8, August~2021}%
{Shell \MakeLowercase{\textit{et al.}}: A Sample Article Using IEEEtran.cls for IEEE Journals}


\maketitle

\begin{abstract}
In conversational machine reading, systems need to interpret natural language rules, answer high-level questions, and ask follow-up clarification questions.
However, existing works assume that all the background knowledge to answer each question is given, which neglects the essential retrieval step in real scenarios.
In this work, we propose and investigate an open-retrieval setting of conversational machine reading.
In the open-retrieval setting, the relevant rule texts are unknown so that a system needs to retrieve question-relevant evidence from a collection of rule texts, and answer users' high-level questions according to multiple retrieved rule texts in a conversational manner.
We propose \modelnameshort, a \modelname\ which extracts conditions in the rule texts through discourse segmentation, conducts multi-passage entailment reasoning to answer user questions directly, or asks clarification follow-up questions to inquiry more information.
On our created \orsharc\ dataset, \modelnameshort\ achieves the \sota\ performance, outperforming existing single-passage conversational machine reading models as well as a new multi-passage conversational machine reading baseline by a large margin.
In addition, we conduct in-depth analyses to provide new insights into this new setting and our model.
\end{abstract}


% abstract 1
% introduction 2 - 5
% related work 5 - 6
% task definition 7 - 9
% model 10 - 15
% experiment 15 - 20

\section{Introduction}

% introduce conversational machine reading
Conversational machine reading aims to answer high-level questions that cannot be answered literally from the text \cite{saeidi-etal-2018-interpretation}.
Instead, questions such as ``May I qualify for VA health care benefits?'' require systems to understand the requirements listed in the rule text and evaluate the user's satisfaction towards these requirements. 
Since users are unaware of the rule text, they cannot provide all their background information needed in a single turn. In this case, systems may keep asking clarification questions on underspecified requirements until the systems can make a final decision (e.g. qualify or not).
This interactive behavior between users and machines receives attention recently \cite{zhong-zettlemoyer-2019-e3,gao-etal-2020-explicit} and has applications in intelligent assistant systems such as Siri, Alexa, and Google Assistant.

\begin{figure}[t!]
\small
\begin{tabular}{p{0.95\columnwidth}}
\hline\hline
\textbf{Retrieved Rule Text 1}: SBA provides loans \textbf{to businesses - not individuals} - so the requirements of eligibility are based on aspects of the business, not the owners. All businesses that are considered for financing under SBA's 7(a) loan program must: \textbf{meet SBA size standards}, be \textbf{for-profit}, \textbf{not already have the internal resources (business or personal) to provide the financing}, and be able to demonstrate repayment.  \\
\textbf{Retrieved Rule Text 2}: You’ll need a statement of National Insurance you’ve paid in the UK to get these benefits - unless you’re claiming Winter Fuel Payments. \\
\textbf{Retrieved Rule Text 3}: 7(a) loans are the most basic and most used type loan of the Small Business Administration's (SBA) business loan programs. It's name comes from section 7(a) of the Small Business Act, which authorizes the agency to provide business loans to \textbf{American small businesses}. The loan program is designed to assist \textbf{for-profit businesses} that are \textbf{not able to get other financing from other resources}. \\
\hline
\textbf{User Scenario}: I am a 34 year old man from the United States who owns their own business. We are an American small business. \\
\textbf{User Question}: Is the 7(a) loan program for me? \\
\textbf{Follow-up} $\mathbf{Q}_1$: Are you a for-profit business? \\
\textbf{Follow-up} $\mathbf{A}_1$: Yes. \\
\textbf{Follow-up} $\mathbf{Q}_2$: Are you able to get financing from other resources? \\
\textbf{Follow-up} $\mathbf{A}_2$: No. \\
\textbf{Final Answer}: Yes. (You can apply the loan.)\\
\hline\hline                
\end{tabular}
\caption{Open retrieval conversational machine reading task. The machine answers the user question by searching for relevant rule texts, reading retrieved noisy rule texts (rule text 2 is irrelevant to the user question), interpreting the user scenario, and keeping asking follow-up questions to clarify the user's background until it concludes a final answer. Question-relevant requirements in the rule texts are bold.}
\label{fig:example}
\end{figure}

% introduce our consideration
However, previous study \cite{saeidi-etal-2018-interpretation} simplify this challenging problem by assuming that the relevant rule text is given before the system interacts with users, which is unrealistic in real intelligent assistant systems.
In this paper, we consider an \textit{Open-Retrieval} setting of conversational machine reading.
In this setting, a system needs to retrieve relevant rule texts according to the user's question, read these rule texts, interact with the user by asking clarification questions, and finally give the answer.
Fig. \ref{fig:example} provides an example for this setting. 
The user asks whether he is suitable for the loan program. The machine retrieves three question-relevant rule texts but only rule text 1 and rule text 3 are truly relevant.
Then it understands the requirements listed in the rule text, finds the user meets ``American small business'' from the user scenario, asks two follow-up questions about ``for-profit business'' and ``not get financing from other resources'', and finally it concludes the answer ``Yes'' to the user's initial question.

% challenges
There are two major challenges for this new setting.
Firstly, existing conversational machine reading datasets such as \sharc\ \cite{saeidi-etal-2018-interpretation} contains incomplete initial questions such as ``Can I receive this benefit?'' because the annotator is given the rule text for question generation in the data collection process. But it brings ambiguity in our open-retrieval setting because we do not know which benefit does the user asks for.
Second, existing conversational machine reading models \cite{zhong-zettlemoyer-2019-e3,gao-etal-2020-explicit,gao-etal-2020-discern,lawrence-etal-2019-attending} ask and answer conversational questions with the single gold rule text but it is not the case for the open-retrieval setting.
Models need to retrieve multiple relevant rule texts and read and reason over the noisy rule texts (e.g. the retrieved rule text 2 in Fig. \ref{fig:example} is irrelevant to the question).

% dataset, model, our solution to the challenge
We tackle the first challenge by reformulating incomplete questions to their complete version according to the gold rule text in \sharc\ dataset, and create an \textit{Open-Retrieval} version of \sharc\ named \orsharc.
For example, the incomplete question ``Can I receive this benefit?'' will be reformulated by specifying the name of the benefit mentioned in the rule text ``Can I receive the benefit of 7(a) Loan Program?''.
Consequently, the ambiguity of the initial user question is resolved.
Furthermore, we use TF-IDF to retrieve multiple relevant rule texts, and propose \modelnameshort: \modelname\ for conversational machine reading over multiple retrieved rule texts.
Firstly, \modelnameshort\ segments rule texts into clause-like elementary discourse units (EDUs).
Then it conducts discourse-aware entailment reasoning according to the user-provided scenario and dialog history to decide which requirements are satisfied in a weakly supervised manner.
Finally, it directly answers the user question with ``Yes/No'' or asks a clarification follow-up question.

% experimental results
We compare \modelnameshort\ with previous best single-passage conversational machine reading models as well as a new RoBERTa-based multi-passage conversational machine reading baseline on our created \orsharc\ dataset. Our results show that \modelnameshort\ can not only make the ``Yes/No/Inquire'' decision more accurate at each dialog turn but also generate better and to-the-point clarification follow-up questions.
In particular, \modelnameshort\ outperforms the previous best model \discern\  \cite{gao-etal-2020-discern} by 8.2\% in micro-averaged decision accuracy and 11.8\% in question generation \Fbleufour.
Moreover, we conduct in-depth analyses on model ablation and configuration to provide insights for the \orsharc\ task.
We show that \modelnameshort\ can perform extremely well on rule texts seen in training but there is still a large room for improvement on unseen rule texts.

% contributions
The main contributions of this work, which are fundamental to significantly push the \sota\ in open-retrieval conversational machine reading, can be summarized as follows:
\begin{itemize}[noitemsep]
    \item We propose a new Open-Retrieval setting of conversational machine reading on natural language rules, and create a dataset \orsharc\ for this setting.
    \item We develop a new model \modelnameshort\ on the open-retrieval conversational machine reading that can jointly read and reason over multiple rule texts for asking and answering conversational questions.
    \item We conduct extensive experiments to show that \modelnameshort\ outperforms existing single-passage models as well as our strong RoBERTa-based multi-passage baseline on the \orsharc\ dataset.
\end{itemize}

\section{Related Work}

\subsection{Conversational Machine Reading}

Machine reading comprehension is a long-standing task in natural language processing and has received a resurgence recently \cite{TASLP9352490,TASLP9165888,TASLP9580589}.
Different from dialog question answering \cite{choi-etal-2018-quac,reddy-etal-2019-coqa,gao-etal-2019-interconnected} which aims to answer factoid questions for information gathering in a conversation and open-domain dialog generation \cite{gao-etal-2020-dialogue-generation} which makes responses without any restriction, conversational machine reading on natural language rules \cite{saeidi-etal-2018-interpretation} require machines to interpret a set of complex requirements in the rule text, ask clarification question to fill the information gap, and make a final decision as the answer does not literally appear in the rule text.
This major difference motivates a series of work focusing on how to improve the joint understanding of the dialog and rule text instead of matching the conversational question to the text in conversational question answering.
Initial methods simply use pretrained models like BERT \cite{devlin-etal-2019-bert} for answer classification without considering satisfaction states of each rule listed in the rule text \cite{verma-etal-2020-neural,lawrence-etal-2019-attending}. 
Zhong and Zettlemoyer \cite{zhong-zettlemoyer-2019-e3} propose to firstly identify all rule text spans, infer the satisfaction of each rule span according to the textual overlap with dialog history for decision making.
Recent works \cite{gao-etal-2020-explicit,gao-etal-2020-discern} focuses on modeling the entailment-oriented reasoning process in conversational machine reading for decision making.
They parse the rule text into several conditions, and explicitly track the entailment state (satisfied or not) for each condition for the final decision making.
Different from all previous work, we consider a new open-retrieval setting where the ground truth rule text is not provided.
Consequently, our \modelnameshort\ retrieves several rule texts and our work focuses on \textit{multi-passage} conversational machine reading which has never been studied.

\subsection{Open Domain Question Answering}
Open-Domain Question Answering is a long-standing problem in NLP \cite{Voorhees1999TheTQ, moldovan-etal-2000-structure, brill-etal-2002-analysis, Ferrucci2010BuildingWA}. 
Different from reading comprehension \cite{rajpurkar-etal-2016-squad}, supporting passages (evidence) are not provided as inputs in open-domain QA.
Instead, all questions are answered using a collection of documents such as Wikipedia pages.
Most recent open-domain QA systems focus on answering factoid questions through a two-stage retriever-reader approach \cite{chen-etal-2017-reading,wang-etal-2019-multi,Gao2020AnsweringAQ}: a retriever module retrieves dozens of question-relevant passages from a large collection of passages using BM25 \cite{lee-etal-2019-latent} or dense representations \cite{karpukhin-etal-2020-dense}, then a reader module finds the answer from the retrieved rule texts.
Chen et al. \cite{10.1145/3397271.3401110} firstly introduces an open-domain setting for answering factoid conversational questions in QuAC \cite{choi-etal-2018-quac}.
They propose ORConvQA, which firstly retrieves evidence from a large collection of passages, reformulates the question according to the dialog history, and extracts the most possible span as the answer from retrieved rule texts. 
However, there are two major differences between ORConvQA and our work:
1) They focus on answering open-domain factoid conversational questions but our work addresses the challenges in answering high-level questions in conversational machine reading;
2) Their proposed model ORConvQA follows the recent reranking-and-extraction approach in open-domain \textit{factoid} question answering \cite{wang-etal-2019-multi} while our \modelnameshort\ aims to answer \textit{high-level} questions in an entailment-driven manner.

\section{Task Definition}

\subsection{\orsharc\ Setup}
Fig. \ref{fig:example} depicts the \orsharc\ task.
The user describes the scenario $s$ and posts the initial question $q$.
The system needs to search in its knowledge base (a large collection of rule texts $R$), and retrieve relevant rule texts $r_1, r_2, ..., r_m$ using the user scenario and question.
Then the conversation starts. At each dialog turn, the input is retrieved rule texts $r_1, r_2, ..., r_m$, a user scenario $s$, user question $q$, and previous $k ~(\geq 0)$ turns dialog history $(q_1, a_1), ..., (q_k, a_k)$ (if any).
\orsharc\ has two subtasks: 1) \textbf{Decision Making}: the system needs to make a decision among ``Yes'', ``No, ``Inquire'' where ``Yes'' \& ``No'' directly answers the user's initial question while ``'inquire'' indicates the system still needs to gather further information from the user.
2) \textbf{Question Generation}: The ``inquire'' decision will let the system generate a follow-up question on the underspecified information for clarification.

\begin{figure*}[!t]
    \centering
    \footnotesize
    \resizebox{1.0\textwidth}{!}{
    \begin{tabular}{p{0.4\textwidth}p{0.2\textwidth}p{0.4\textwidth}}
    \toprule
        Rule Text & Original Question & Rewritten Question \\
    \midrule
        The Fisheries Finance Program (FFP) is a direct government loan program that provides long term financing for the cost of construction or reconstruction of fishing vessels, fisheries facilities, aquacultural facilities and individual fishing quota in the Northwest Halibut/Sablefish and Alaskan Crab Fisheries.  &	\multirow{5}{*}{Does this loan meet my needs?}  &	\multirow{5}{*}{Does the loan from Fisheries Finance Program meet my needs?} \\
    \specialrule{0em}{1pt}{1pt}
    \hdashline
    \specialrule{0em}{1pt}{1pt}
        Loans are for 33 years at 1 percent interest. Grants may cover up to 90 percent of development costs. The balance may be a Farm Labor Housing Program loan. &	\multirow{3}{*}{Does this loan meet my needs?} &	\multirow{3}{*}{Does this Farm Labor Housing Program loan meet my needs?} \\
    \specialrule{0em}{1pt}{1pt}
    \hdashline
    \specialrule{0em}{1pt}{1pt}
        Farm Storage Facility Loans (FSFL) are provided to encourage the construction of on-farm storage and handling facilities for eligible commodities. Eligible commodities include: corn, grain sorghum, oats, wheat, barley, rice, soybeans, peanuts .... &	\multirow{4}{*}{Does this loan meet my needs?} &	\multirow{4}{*}{Does this Farm Storage Facility loan meet my needs?} \\
    \bottomrule
    \end{tabular}
    }
    \caption{
        Examples of rule texts and original questions in the ShARC dataset, and the rewritten questions for the open-retrieval setting in the OR-ShARC dataset.
    }
    % \vspace{-0.1in}
    \label{fig:qr_case}
\end{figure*}

\subsection{\orsharc\ Dataset}\label{sec:dataset}
The \sharc\ \cite{saeidi-etal-2018-interpretation} dataset is designed for conversational machine reading.
During the annotation stage, the annotator describes his scenario and asks a high-level question on the given rule text which cannot be answered in a single turn. 
Instead, another annotator acts as the machine to ask follow-up questions whose answer is necessary to answer the original question. 
We reuse the user question, scenario, and constructed dialog including follow-up questions and answers for our open-retrieval setting.
The difference is that we remove the gold rule text for each sample.
Instead, we collect all rule texts used in the \sharc\ dataset as the knowledge base which is accessible for all training and evaluation samples. 
% \footnote{Different from the open-domain question answering task, we cannot extract a huge collection of passages directly from websites such as Wikipedia because every rule text contains an inherent logic structure that requires careful extraction. Therefore, we build our collection of passages directly on the ground truth rule text extracted from \sharc\ annotators.}
Since the test set of \sharc\ is not public, we manually split the train, validation, and test set. For the validation and test set, around half of the samples ask questions on rule texts used in training (seen) while the remaining of them contain questions on unseen (new) rule texts.
We design the seen and unseen splits for the validation and test set because it mimics the real usage scenario -- users may ask questions about rule text which we have training data (dialog history, scenario) as well as completely newly added rule text.
The dataset statistics is shown in Table \ref{tab:dataset}.

\begin{table}[!t]
	\centering
% 	\small
    \caption{
	   The statistics of \orsharc\ dataset. 
	}\label{tab:dataset}
% 	\resizebox{0.7\textwidth}{!}{
	\begin{tabular}{lc}
    \toprule
        \# Rule Texts & 651   \\
        \# Avg. Rule Text Length & 38.5 \\
        \# Train Samples & 17936  \\
        \# Dev Samples 	& 1105 \\
        % ~~ w/ seen passages & 500 \\
        % ~~ w/ unseen passages & 605 \\
        \# Test Samples & 2373 \\
        % ~~ w/ seen passages & 1000 \\
        % ~~ w/ unseen passages & 1373 \\
        \# Avg. Dialog Turns	& 1.4 \\
    \bottomrule
    \end{tabular}
    % }
\end{table}

Another challenge in our open-retrieval setting is that user questions must be complete and meaningful so that the retrieval system can search for relevant rule texts, which is not the case for \sharc\ dataset.
In \sharc\ dataset, annotators may not provide the complete information needs in their initial questions because they have already seen the rule texts.
In other words, the information need is jointly defined by the initial question and its corresponding rule text in the \sharc\ dataset.
As shown in Fig. \ref{fig:qr_case}, questions such as ``Does this loan meet my needs?'' have different meanings when they are paired with rule texts on different benefits.
Since the gold rule text is unknown in our open-retrieval setting, the real information need may not be captured successfully.
We rewrite all questions in the \sharc\ dataset when the information need cannot be solely determined by the initial user questions.
In the above case, the loan program in the question ``Does this loan meet my needs?'' will be further specified according to the rule texts. In total, 12.2\% questions in the ShARC dataset are rewritten according to the above protocol.
Our collected dataset \orsharc\ is released at \url{https://github.com/Yifan-Gao/open_retrieval_conversational_machine_reading}.

\subsection{Evaluation Metrics}
\paragraph{Retrieval Task.} 
We evaluate the Top-K ($K=1,2,5,10,20$) recall of the gold rule text for the retrieval task.
\paragraph{Decision Making Task.} 
For the decision-making subtask. we use macro- and micro-accuracy of three classes ``Yes'', ``No'' and ``Inquire'' as evaluation metrics.
\paragraph{Question Generation Task.} 
We compute the BLEU score \cite{papineni-etal-2002-bleu} between the generated questions and the ground truth questions.
However, different models may have a different set of ``Inquire'' decisions and therefore predicts a different set of follow-up questions, which make the results non-comparable.
Previous work either evaluate the BLEU score only when both the ground truth decision and the predicted decision are ``Inquire'' \cite{saeidi-etal-2018-interpretation} or asks the model to generate a question when the ground truth decision is ``Inquire'' \cite{gao-etal-2020-explicit}.
We think both ways of evaluation are sub-optimal because the former only evaluates the intersection of ``Inquire'' decisions between prediction and ground truth questions without penalizing the model with less accurate ``Inquire'' predictions while the latter approach requires separate prediction on the oracle question generation task.
Here we propose a new metric \Fbleu\ for question generation.
We compute the precision of BLEU for question generation when the \textit{predicted decision} is ``Inquire'',
\begin{align}
    precision_{\text{BLEU}} = \frac{\sum_{i=0}^{M} \text{BLEU}(y_i, \hat{y}_i)}{M},
\end{align}
where $y_i$ is the predicted question, $\hat{y}_i$ is the corresponding ground truth prediction when the model makes an ``Inquire'' decision ($\hat{y}_i$ is not necessarily a follow-up question), and $M$ is the total number of ``Inquire'' decisions from the model.
The recall of BLEU is computed in a similar way when the \textit{ground truth decision} is ``Inquire'',
\begin{align}
    recall_{\text{BLEU}} = \frac{\sum_{i=0}^{N} \text{BLEU}(y_i, \hat{y}_i)}{N},
\end{align}
where $N$ is the total number of ``Inquire'' decision from the ground truth annotation, and the model prediction $y_i$ here is not necessarily a follow-up question.
We finally calculate F1 by combining the precision and recall,
\begin{align}
    \text{F1}_{\text{BLEU}} = \frac{2 \times precision_{\text{BLEU}} \times recall_{\text{BLEU}}}{precision_{\text{BLEU}} + recall_{\text{BLEU}}}.
\end{align}

\begin{figure*}[t!]
\centering
\includegraphics[width=0.6\textwidth]{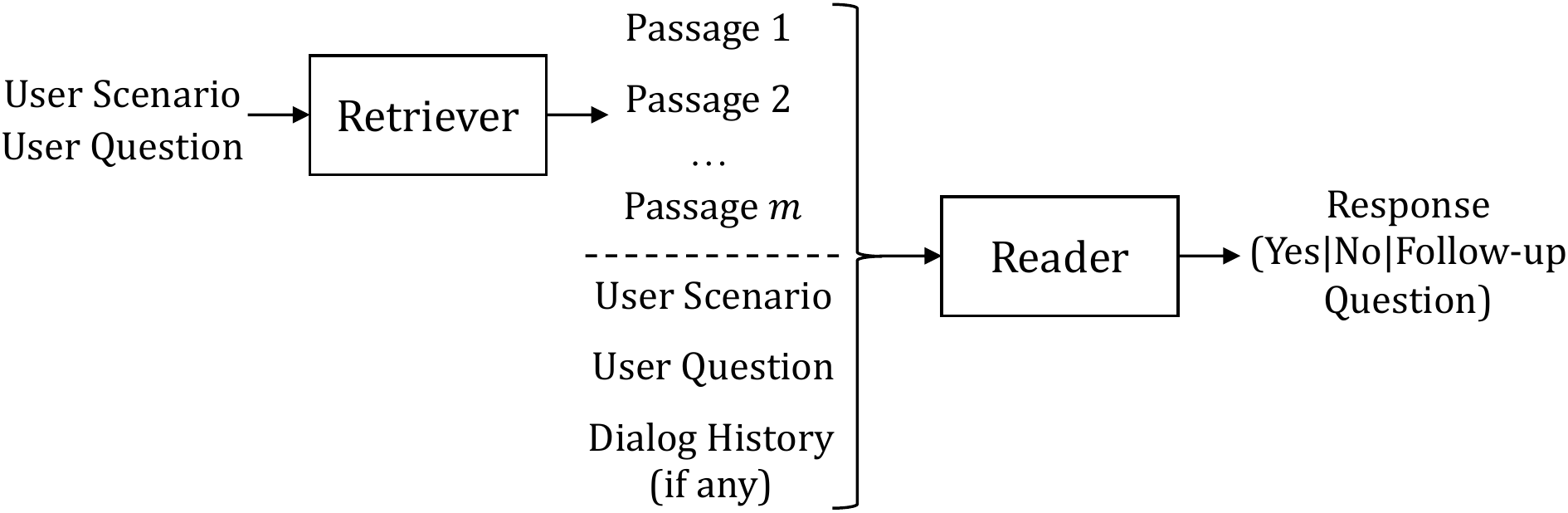}
\caption{The overall diagram of our proposed \modelnameshort. \modelnameshort\ first retrieves several relevant passages (rule texts) using the user provided scenario and question (Section \ref{sec:retriever}). Then, for every dialog turn, taking the retrieved rule texts, user question, user scenario, and a history of previous follow-up questions and answers (if any) as inputs, \modelnameshort predict the answer to the question (``Yes'' or ``No'') or, if needed, generate a follow-up question whose answer is necessary to answer the original question (Section \ref{sec:reader}).}
\label{fig:framework}
\end{figure*}

\section{Framework}

Our framework answers the user question through a retriever-reader process shown in Fig. \ref{fig:framework}:

\begin{enumerate}[noitemsep]
    \item Taking the concatenation of user question and scenario as query, the TF-IDF retriever gets $m$ most relevant rule texts for the reader.
    \item Our reader \modelnameshort\ initiates a turn-by-turn conversation with the user. At each dialog turn,
    taking the retrieved rule texts and the user-provided information including the user question, scenario, and dialog history as inputs, \modelnameshort\ predicts ``Yes'' or ``No'' as the answer or asks a follow-up question for clarification.
\end{enumerate}

\subsection{TF-IDF Retriever}\label{sec:retriever}
The passage retriever module narrows the search space so that the reader module can focus on reading possibly relevant passages (``passage'' and ``rule text'' are used interchangeably in the rest of this paper).
We concatenate the user question and user scenario as the query, and use a simple inverted index lookup followed by term vector model scoring to retrieve relevant rule texts.
Rule texts and queries are compared as TF-IDF weighted bag-of-words vectors.
We further take the local word order into account with bigram features which is proven to be effective in \cite{chen-etal-2017-reading}.
We retrieve 20 rule texts for each query, which is used by the reader module.

\subsection{Reader: \modelnameshort}\label{sec:reader}

We propose \modelnameshort: \modelname\ as the reader, and it reads and answers questions via a three-step approach:
Firstly \modelnameshort\ uses discourse segmentation to parse each retrieved rule text into individual conditions (Section \ref{sec:reader-discourse}).
Secondly, \modelnameshort\ reads the parsed conditions and the user-provided information including the user question, scenario, and dialog history, and makes the decision among ``Yes'', ``No'', and ``Inquire'' (Section \ref{sec:reader-dm}).
Finally, if the decision is ``Inquire'', \modelnameshort\ generates a follow-up question to clarify the underspecified condition in the retrieved rule texts (Section \ref{sec:reader-qg}).

\subsubsection{Discourse-aware Rule Segmentation}\label{sec:reader-discourse}
One challenge in understanding rule texts is that the rule texts have complex logical structures.
For example, the rule text ``If a worker has taken more leave than they’re entitled to, their employer must not take money from their final pay unless it’s been agreed beforehand in writing.'' actually contains two conditions (``If a worker has taken more leave than they’re entitled to'', ``unless it’s been agreed beforehand in writing'') and one outcome (``their employer must not take money from their final pay'').
Without successful identification of these conditions, the reader cannot ensure the satisfaction state of each condition and reason out the correct decision.
Here we use discourse segmentation in discourse parsing to extract these conditions.
In discourse parsing \cite{mann1988rhetorical}, texts are firstly split into elementary discourse units (EDUs). Each EDU is a clause-like span in the text.
We utilize an off-the-shelf discourse segmenter \cite{DBLP:conf/ijcai/LiSJ18} for rule segmentation, which is a pre-trained pointer model that achieves near human-level accuracy (92.2\% F-score) on the standard benchmark test set.

\begin{figure*}[t!]
\centering
\includegraphics[width=1.0\textwidth]{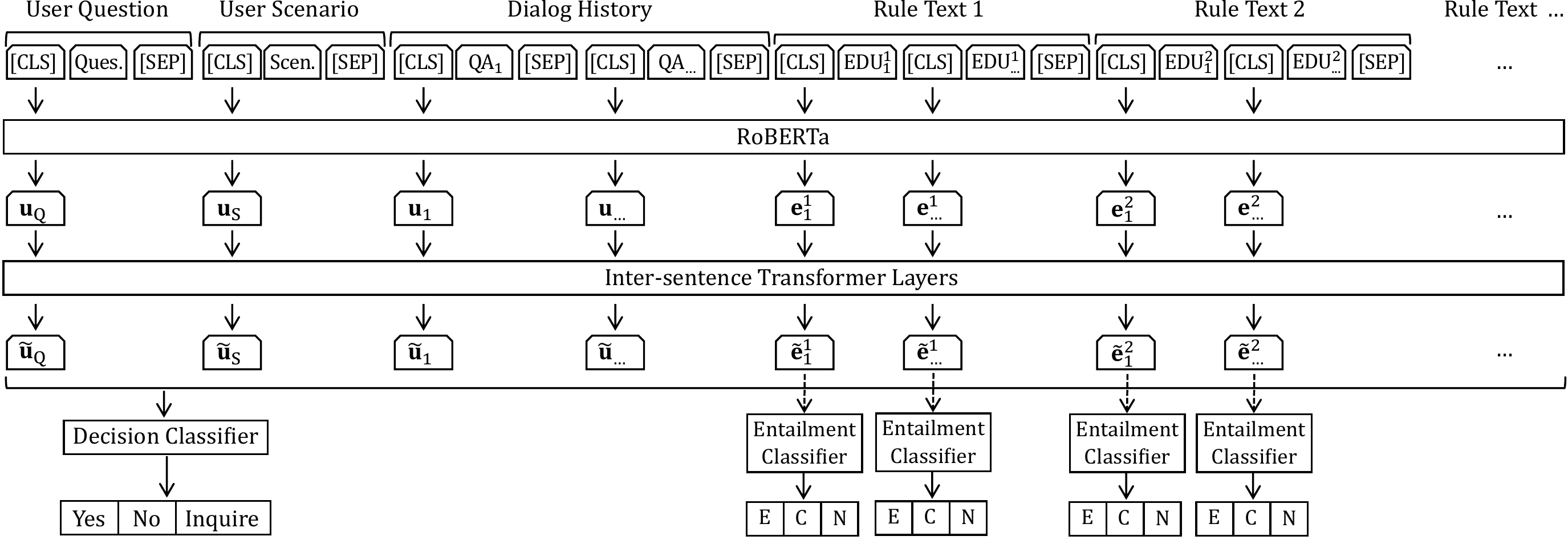}
\caption{The decision making module of our proposed \modelnameshort\ (Section \ref{sec:reader-dm}).
The module makes a decision among ``Yes'', ``No'', and ``Inquire'', as well as predicts the entailment states for every elementary discourse unit (EDU) among ``Entailment'' (E), ``Contradiction'' (C), and ``Neutral'' (N).}
\label{fig:model-dm}
\end{figure*}

\subsubsection{Entailment-driven Decision Making}\label{sec:reader-dm}

\paragraph{Encoding}
As shown in Fig. \ref{fig:model-dm}, 
we concatenate the user question, user scenario, a history of previous follow-up questions
and answers, and the segmented EDUs from the retrieved rule texts into a sequence, and use RoBERTa \cite{liu2019roberta} for encoding.
Additionally, we insert a \texttt{[CLS]} token at the start and append a \texttt{[SEP]} token at the end of all individual sequences to get the sentence-level representations.
Each \texttt{[CLS]} represents the sequence that follows it.
Finally, we receive the representations of user question $\mathbf u_Q$, user scenario $\mathbf u_S$, $H$ turns of follow-up question and answers $\mathbf u_1, ..., \mathbf u_H$, and segmented conditions (EDUs) from $m$ retrieved text $\mathbf e^1_1, \mathbf e^1_2, ..., \mathbf e^1_N; ..., \mathbf e^m_1, \mathbf e^m_2, ..., \mathbf e^m_N$.
For each sample in the dataset, we append as many complete rule texts as possible under the limit of the maximum input sequence length of RoBERTa (denoted as $\tilde{m}$ rule texts).
In training, if the ground truth rule text does not appear in the top-$\tilde{m}$ retrieved rule text, we manually insert the gold rule text right after the dialog history in the input sequence.
All these vectorized representations are of $d$ dimensions (768 for RoBERTa-base).

We further utilize an inter-sentence transformer encoder \cite{transformer} to learn the sentence-level interactions.
The self-attention in the $L$-layer transformer encoder makes conditions (sentence-level representations) attend each other.
The representations encoded by the inter-sentence transformer are denoted by adding a tilde over the original representations, i.e., $\mathbf u_Q$ becomes $\tilde{\mathbf{u}}_Q$.

\paragraph{Entailment Prediction}

One key step for correct decision-making is the successful determination of the satisfaction state of every condition in rule texts.
Here we propose to formulate this into a multi-sentence entailment task:
Given a sequence of conditions (EDUs) across multiple retrieved rule texts and a sequence of user-provided information, a system needs to predict \textsc{Entailment}, \textsc{Contradiction} or \textsc{Neutral} for each condition.
The condition and user information correspond to premise and hypothesis in the textual entailment task.
\textsc{Neutral} indicates that the condition has not been mentioned from the user information, which is extremely useful in the multi-passage setting because all conditions in the irrelevant rule texts should have this state.
Formally, the $i$-th condition in the $j$-th retrieved rule text $\tilde{\mathbf{e}}^j_i$ goes through a linear layer to receive its entailment state:
\begin{align}
    \mathbf c^j_i &= \mathbf W_c \tilde{\mathbf{e}}^j_i + \mathbf b_c \in \real{3}.
\end{align}

As there are no gold entailment labels for individual conditions in the \sharc\ dataset, we heuristically generate noisy supervision signals similar to \cite{gao-etal-2020-explicit} as follows:
Because each follow-up question-answer pair in the dialog history asks about an individual condition in the rule text, we match them to the segmented condition (EDU) in the rule text which has the minimum edit distance.
Then the entailment states are labelled according to the matching results:
\begin{itemize}
    \item Conditions matched by any follow-up QA pair: If the answer of the follow-up question is \texttt{Yes}, we label the entailment state of the condition as \texttt{Entailment}. Otherwise, we label the state as \texttt{Contradiction} if the answer is \texttt{No}.
    \item Unmentioned conditions: If a condition is not mentioned by any QA pair, we label it as \texttt{Neutral}.
\end{itemize}
Let $r$ indicate the correct entailment state. 
The entailment prediction is supervised by the following cross-entropy loss using our generated noisy labels, normalized by total number of $K$ conditions in a batch:
\begin{align}\label{eqn:loss-entailment}
    \mathcal{L}_{\text{entail}} = - \frac{1}{K}\sum_{(i,j)} ~ \log~\text{softmax}(\mathbf c^j_i)_r
\end{align}

\paragraph{Decision Making}
Given encoded sentence-level user information $\tilde{\mathbf{u}}_Q, \tilde{\mathbf{u}}_S, \tilde{\mathbf{u}}_1, ..., \tilde{\mathbf{u}}_H$ and rule conditions $\tilde{\mathbf{e}}^1_1, \tilde{\mathbf{e}}^1_2, ..., \tilde{\mathbf{e}}^1_N; ..., \tilde{\mathbf{e}}^{\tilde{m}}_1, \tilde{\mathbf{e}}^{\tilde{m}}_2, ..., \tilde{\mathbf{e}}^{\tilde{m}}_N$, the decision making module predicts ``Yes'', ``No'', or ``Inquire'' for the current dialog turn.
Here we do not distinguish the sequence from the user information or rule texts, and denote all of them into ${\mathbf{h}}_1, {\mathbf{h}}_2, ..., {\mathbf{h}}_n$. We start by computing a summary $z$ of the rule texts and user information using self-attention:
\begin{align}
    \alpha_i &= \mathbf w_{\alpha}^\top {\mathbf{h}}_i + b_\alpha \in \real{1} \label{eqn:alpha} \\
    \tilde{\alpha}_i &= \text{softmax}(\mathbf{\alpha})_i \in [0,1]  \\
    \mathbf g &= \sum_i \tilde{\alpha}_i {\mathbf{h}}_i \in \real{d} \label{eqn:summary} \\
    \mathbf z &= \mathbf W_z \mathbf g + \mathbf b_z \in \real{3}
\end{align}
where $\alpha_i$ is the self-attention weight. 
$\mathbf z \in \real{3}$ contains the predicted scores for all three possible decisions ``Yes'', ``No'' and ``Inquire''.
Let $l$ indicate the correct decision, $\mathbf z$ is supervised by the following cross-entropy loss:
\begin{align}\label{eqn:loss-dm}
    \mathcal{L}_{\text{dec}} = -\log~\text{softmax}(\mathbf z)_l
\end{align}

\noindent The overall loss for the entailment-driven decision making module is the weighted-sum of decision loss and entailment prediction loss:
\begin{align}
    \mathcal{L_\text{dm}} = \mathcal{L}_{\text{dec}} + \lambda \mathcal{L}_{\text{entail}}, \label{eq:totalloss}
\end{align}
where $\lambda$ is a hyperparameter tuned on the validation set.

\subsubsection{Follow-up Question Generation}\label{sec:reader-qg}

If the predicted decision is ``Inquire'', a follow-up question is required as its answer is necessary to answer the original question.
Recent approaches in question generation usually identify the to-be-asked context first, and generate the complete question \cite{li-etal-2019-improving-question,ijcai2019-0690}.
Similarly, we design an extraction-and-rewrite approach which firstly extracts an underspecified span within the rule texts and then rewrites it into a well-formed question.

\paragraph{Span Extraction}
Similar to the encoding step in the decision-making module (Section \ref{sec:reader-dm}), we concatenate the user information and rule texts into a sequence, and use RoBERTa for encoding.
The only difference here is we do not further segment sentences in the rule texts into EDUs because we find sometimes the follow-up question contains tokens across multiple EDUs.
Let [$\mathbf{t}_{1,1}$, ..., $\mathbf{t}_{1,s_1}$; $\mathbf{t}_{2,1}$, ..., $\mathbf{t}_{2,s_2}$; ...; $\mathbf{t}_{N,1}$, ..., $\mathbf{t}_{N,s_N}$] be the encoded vectors for tokens from $N$ rule sentences in retrieved rule texts, we follow the BERTQA approach \cite{devlin-etal-2019-bert} to learn a start vector $\mathbf w_s \in \real{d}$ and an end vector $\mathbf w_e \in \real{d}$ to locate the start and end positions, under the restriction that the start and end positions must belong to the same rule sentence:
\begin{align}\label{eqn:span}
    \text{Span} = \argmax_{i,j,k} (\mathbf{w}_s^\top \mathbf{t}_{k,i} + \mathbf{w}_e^\top \mathbf{t}_{k,j})
\end{align}
where $i,j$ denote the start and end positions of the selected span, and $k$ is the sentence which the span belongs to. 
Similar to the training in the decision-making module, we include the ground truth rule text if it does not appear in the top-$\tilde{m}$ retrieved texts.
We train the span extraction model using the log-likelihood of the gold start and and positions.
We create the noisy labels of spans' starts and ends by selecting the span which has the minimum edit distance with the to-be-asked follow-up question.

\paragraph{BART-based Question Rewriting}
Given the to-be-asked context (extracted span), we find the rule text it belongs to, and concatenate the extracted span with the rule text into one sequence as ``\texttt{[CLS]} extracted-span \texttt{[SEP]} rule-text \texttt{[SEP]}''.
Then we use BART \cite{lewis-etal-2020-bart}, a pre-trained seq2seq language model, to rewrite the sequence into a follow-up question.
During training, the BART model takes the ground truth rule text and our created noisy spans as inputs, and the target is the corresponding follow-up question.

\section{Experiments}

\subsection{Implementation Details}
For the decision making module (Section \ref{sec:reader-dm}), we finetune RoBERTa-base model \cite{wolf-etal-2020-transformers} with Adam \cite{adam} optimizer for 10 epochs with a learning rate of 3e-5 and a batch size of 32.
We treat the number of inter-sentence transformer layers $L$ and the loss weight $\lambda$ for entailment prediction as hyperparameters, tune them on the validation set through grid search, and find the best combination is $L=4, \lambda=8$.
For the span extraction module (Section \ref{sec:reader-qg}), we finetune RoBERTa-base model with Adam optimizer for 10 epochs with a learning rate of 5e-5 and a batch size of 32.
For the question rewriting module (Section \ref{sec:reader-qg}), we finetune BART-base model \cite{wolf-etal-2020-transformers} with Adam optimizer for 30 epochs with a learning rate of 1e-4 and a batch size of 32.
These parameters are tuned using the validation set.
We repeat 5 times with different random seeds and report average results and standard deviations.
Code and models are released at \url{https://github.com/Yifan-Gao/open_retrieval_conversational_machine_reading}.

\subsection{Baselines}
Since we are the first to investigate this open-retrieval conversational machine reading setting, we replicate several previous works on \sharc\ conversational machine reading dataset including \ethree\ \cite{zhong-zettlemoyer-2019-e3}, \emt\ \cite{gao-etal-2020-explicit}, \discern\ \cite{gao-etal-2020-discern}, and adapt them to our open-retrieval setting: 
We train them in the same way as conversational machine reading using the gold rule text on the \orsharc\ training set, and use the top-1 retrieved rule text from our TF-IDF retriever for prediction and evaluation.
We also propose a multi-passage baseline \mprbt\ which concatenate all sources of inputs including user question, scenario, dialog history, and multiple retrieved rule texts in a sequence, insert a \texttt{[CLS]} token at first, and make the decision prediction using that \texttt{[CLS]} token through a linear layer.
\mprbt\ shares the same TF-IDF retriever (Section \ref{sec:retriever}) and question generation module (Section \ref{sec:reader-qg}) as our proposed \modelnameshort.

\begin{table}[!t]
	\centering
% 	\small
	\caption{
	   TF-IDF retriever recall at different top K passages on \orsharc.
	}\label{tab:retrieval}
% 	\resizebox{0.5\columnwidth}{!}{
	\begin{tabular}{cccc}
    \toprule
        TOP K & Train &  Dev & Test  \\
    \midrule
        1	& 59.9 & 53.8 &	66.9 \\
        2	& 71.6 & 67.4 &	76.8 \\
        % 3	& 77.3 & 74.4 &	83.0 \\
        % 4	& 80.9 & 80.4 &	87.9 \\
        5	& 83.8 & 83.4 &	90.3 \\
        10	& 90.4 & 94.0 & 94.0 \\
        20	& 94.2 & 96.6 &	96.6 \\
    \midrule
        \modelnameshort	& 86.2 & 86.4 &	90.8 \\
    \bottomrule
    \end{tabular}
    % }
\end{table}

\subsection{Retriever Results}
Recall at the different number of top retrieved rule texts on the \orsharc\ dataset is shown in Table \ref{tab:retrieval}.
We observe that a simple TF-IDF retriever performs pretty well on \orsharc. With only 5 passages, over 90\% samples on the test set can contain the ground truth rule text.
Since the number of input passages for \modelnameshort\ is restricted by the maximum sequence length of RoBERTa-base (for example, if the user scenario and dialog history uses too many tokens, then the RoBERTa-base model can only take a relatively fewer number of rule texts), we also show the actual recall of input rule texts for \modelnameshort\ in Table \ref{tab:retrieval}.
On average, the number of input rule texts for \modelnameshort\ is equivalent to 5 $\sim$ 6 retrieved rule texts.

\subsection{Reader Results}

Table \ref{tab:main} shows the decision making (macro- and micro- accuracy) and question generation (\Fbleuone, \Fbleufour) results of \orsharc, and we have the following observations:
\begin{itemize}[noitemsep]
    \item Under the open-retrieval setting, there is a large performance gap for single-passage models compared with their performance on the \sharc\ dataset. For example, \discern\ \cite{gao-etal-2020-discern} has 73.2 micro-accuracy on \sharc\ dataset while it is 67.1 on \orsharc. This is because the top-1 retrieved rule text may not be the ground truth text. Therefore, all single-passage baselines (\ethree, \emt, \discern) suffer from the retrieval error.
    \item Reading multiple rule texts has great benefit in the open-retrieval setting. Without sophisticated model design, \mprbt\ outperforms all single-passage baselines with complicated architectures by a large margin.
    \item With the help of discourse-aware rule segmentation and entailment-oriented reasoning, our proposed \modelnameshort\ achieves the best performance across all evaluation metrics. We conduct an ablation study in Section \ref{sec:ablation} to demonstrate the effectiveness of each component of it.
\end{itemize}

\begin{table*}[!t]
	\centering
% 	\small
    \caption{
	   Decision making and question generation results on the validation and test set of \orsharc. The average results with standard deviation on 5 random seeds are reported.
	}\label{tab:main}
% 	\resizebox{1.0\textwidth}{!}{
	\begin{tabular}{lcccccccccc}
    \toprule
        \multirow{2}{*}{Model} & \multicolumn{2}{c}{\macro} &  \multicolumn{2}{c}{\micro} & \multicolumn{2}{c}{\Fbleuone} & \multicolumn{2}{c}{\Fbleufour}  \\
        \cmidrule(lr){2-3} \cmidrule(lr){4-5} \cmidrule(lr){6-7} \cmidrule(lr){8-9} 
        & {dev} & {test} & {dev} & {test} &{dev} & {test} & {dev} & {test} \\
    \midrule
        \ethree\ \cite{zhong-zettlemoyer-2019-e3} & 
        62.3\scriptsize{$\pm$0.9} & 61.7\scriptsize{$\pm$1.9} & 
        61.8\scriptsize{$\pm$0.9} & 61.4\scriptsize{$\pm$2.2} & 
        29.0\scriptsize{$\pm$1.2} & 31.7\scriptsize{$\pm$0.8} & 
        18.1\scriptsize{$\pm$1.0} & 22.2\scriptsize{$\pm$1.1} \\
        \emt\ \cite{gao-etal-2020-explicit} & 
        66.5\scriptsize{$\pm$1.5} & 64.8\scriptsize{$\pm$0.4} & 
        65.6\scriptsize{$\pm$1.6} & 64.3\scriptsize{$\pm$0.5} & 
        36.8\scriptsize{$\pm$1.1} & 38.5\scriptsize{$\pm$0.5} & 
        32.9\scriptsize{$\pm$1.1} & 30.6\scriptsize{$\pm$0.4}  \\
        \discern\ \cite{gao-etal-2020-discern} & 
        66.7\scriptsize{$\pm$1.8} & 67.1\scriptsize{$\pm$1.2} & 
        66.0\scriptsize{$\pm$1.6} & 66.7\scriptsize{$\pm$1.1} & 
        36.3\scriptsize{$\pm$1.9} & 36.7\scriptsize{$\pm$1.4} & 
        28.4\scriptsize{$\pm$2.1} & 28.6\scriptsize{$\pm$1.2}  \\
        \mprbt\  & 
        73.1\scriptsize{$\pm$1.6} & 70.1\scriptsize{$\pm$1.4} & 
        73.0\scriptsize{$\pm$1.7} & 70.4\scriptsize{$\pm$1.5} & 
        45.9\scriptsize{$\pm$1.1} & 40.1\scriptsize{$\pm$1.6} & 
        40.0\scriptsize{$\pm$0.9} & 34.3\scriptsize{$\pm$1.5}  \\
        \modelnameshort\ & 
        \textbf{78.8}\scriptsize{$\pm$0.6} & \textbf{75.3}\scriptsize{$\pm$0.9} & 
        \textbf{78.4}\scriptsize{$\pm$0.5} & \textbf{75.2}\scriptsize{$\pm$1.0} & 
        \textbf{49.9}\scriptsize{$\pm$0.8} & \textbf{47.1}\scriptsize{$\pm$1.7} & 
        \textbf{42.7}\scriptsize{$\pm$0.8} & \textbf{40.4}\scriptsize{$\pm$1.8}  \\
    \bottomrule
    \end{tabular}
    % }
	
    % \vspace{-0.1in}
\end{table*}

\begin{table}[!t]
	\centering
% 	\small
    \caption{
	   Class-wise decision making accuracy among ``Yes'', ``No'' and ``Inquire'' on the validation and test set of \orsharc. 
	}\label{tab:class-wise}
% 	\resizebox{0.6\textwidth}{!}{
	\begin{tabular}{lcccccccc}
    \toprule
        \multirow{2}{*}{Model} & \multicolumn{2}{c}{Yes} &  \multicolumn{2}{c}{No} & \multicolumn{2}{c}{Inquire}  \\
        \cmidrule(lr){2-3} \cmidrule(lr){4-5} \cmidrule(lr){6-7} 
        & {dev} & {test} & {dev} & {test} &{dev} & {test}  \\
    \midrule
        \ethree\ \cite{zhong-zettlemoyer-2019-e3} & 
        58.5 & 58.5 & 
        61.8 & 60.1 & 
        66.5 & 66.4 \\
        \emt\ \cite{gao-etal-2020-explicit} & 
        56.9 & 55.4 & 
        68.6 & 65.6 & 
        74.0 & 73.6 \\
        \discern\ \cite{gao-etal-2020-discern} & 
        61.7 & 65.8 & 
        61.1 & 61.8 & 
        77.3 & 73.6 \\
        \mprbt\  & 
        68.9 & 72.6 & 
        \textbf{80.8} & \textbf{74.2} & 
        69.5 & 63.4 \\
        \modelnameshort\ & 
        \textbf{73.9} & \textbf{76.4} & 
        \textbf{80.8} & 72.2 & 
        \textbf{81.7} & \textbf{77.4} \\
    \bottomrule
    \end{tabular}
    % }
\end{table}

Table \ref{tab:class-wise} presents a class-wise decision-making accuracy.
We find that \modelnameshort\ have far better predictions in most prediction classes, especially on the ``Inquire'' class.
This is because \modelnameshort\ predicts the decision classes based on the entailment states of all conditions from multiple passages.

\begin{table}[!t]
	\centering
% 	\small
	\caption{
	   Ablation Study of \modelnameshort\ on the validation and test set of \orsharc. 
	}\label{tab:ablation}
	\resizebox{1.0\columnwidth}{!}{
	\begin{tabular}{lcccccccccc}
    \toprule
        \multirow{2}{*}{Model} & \multicolumn{2}{c}{\macro} &  \multicolumn{2}{c}{\micro}  \\
        \cmidrule(lr){2-3} \cmidrule(lr){4-5}  
        & {dev} & {test} & {dev} & {test} \\
    \midrule
        \modelnameshort\ & 
        {78.8}\scriptsize{$\pm$0.6} & {75.3}\scriptsize{$\pm$0.9} & 
        {78.4}\scriptsize{$\pm$0.5} & {75.2}\scriptsize{$\pm$1.0} \\
        ~~ w/o Discourse Segmentation & 
        78.1\scriptsize{$\pm$1.7} & 74.6\scriptsize{$\pm$1.0} & 
        77.7\scriptsize{$\pm$1.6} & 74.4\scriptsize{$\pm$1.0} \\
        ~~ w/o Transformer Layers & 
        77.4\scriptsize{$\pm$1.0} & 74.2\scriptsize{$\pm$0.8} & 
        77.1\scriptsize{$\pm$1.0} & 74.2\scriptsize{$\pm$0.8} \\
        ~~ w/o Entailment Reasoning  & 
        73.3\scriptsize{$\pm$0.9} & 71.4\scriptsize{$\pm$1.5} & 
        73.2\scriptsize{$\pm$0.8} & 71.6\scriptsize{$\pm$1.5} \\
        % ~~ w/ \discern\ \cite{gao-etal-2020-discern} & 
        % {78.5}\scriptsize{$\pm$1.0} & {74.9}\scriptsize{$\pm$0.6} & 
        % {78.2}\scriptsize{$\pm$0.9} & {74.8}\scriptsize{$\pm$0.6} \\
    \bottomrule
    \end{tabular}
    }
	
    % \vspace{-0.1in}
\end{table}

\subsection{Ablation Study}\label{sec:ablation}
We conduct an ablation study to investigate the effectiveness of each component of \modelnameshort\ in the decision-making task. Results are shown in Table \ref{tab:ablation} and our observations are as follows:

\begin{itemize}[noitemsep]
    \item We replace the discourse segmentation with simple sentence splitting to examine the necessity of using discourse. 
    Results drop 0.7\% on macro-accuracy and 0.8\% on micro-accuracy for the test set (``w/o Discourse Segmentation''). 
    This shows the discourse segmentation does help for decision predictions, especially when the rule sentence includes multiple clauses.
    \item We find the inter-sentence transformer layers do help the decision prediction process as there is a 1.1\% performance drop on the test set macro-accuracy if we remove them (``w/o Transformer Layers''). 
    This tells us that the token-level multi-head self-attention in the RoBERTa-base model and the sentence-level self-attention in our added inter-sentence transformer layers are complementary to each other.
    \item The model performs much worse without the entailment supervision (``w/o Entailment Reasoning''), leading to a 3.9\% drop on the test set macro-accuracy.
    It implies the entailment reasoning is essential for the conversational machine reading which is the key difference between \orsharc\ and open-retrieval conversational question answering \cite{10.1145/3397271.3401110}.
    % \item We investigate the \discern-like \cite{gao-etal-2020-discern} architecture in the multi-passage setting (``w/ \discern'').
    % We observe that their proposed entailment vector and making the decision based on the entailment predictions become less effective in the multi-passage scenario.
    % Our \modelnameshort\ has a simpler decision-making module but still achieves slightly better results.
\end{itemize}

\begin{table*}[!t]
	\centering
% 	\small
    \caption{
	   Results on the seen split and unseen split of \orsharc. 
	}\label{tab:snippet-analysis}
% 	\resizebox{1.0\textwidth}{!}{
	\begin{tabular}{lcccccccccc}
    \toprule
        \multirow{2}{*}{Dataset} & \multicolumn{2}{c}{\macro} &  \multicolumn{2}{c}{\micro} & \multicolumn{2}{c}{\Fbleuone} & \multicolumn{2}{c}{\Fbleufour}  \\
        \cmidrule(lr){2-3} \cmidrule(lr){4-5} \cmidrule(lr){6-7} \cmidrule(lr){8-9} 
        & {dev} & {test} & {dev} & {test} &{dev} & {test} & {dev} & {test} \\
    \midrule
        Full Dataset & 
        {78.8}\scriptsize{$\pm$0.6} & {75.3}\scriptsize{$\pm$0.9} & 
        {78.4}\scriptsize{$\pm$0.5} & {75.2}\scriptsize{$\pm$1.0} & 
        {49.9}\scriptsize{$\pm$0.8} & {47.1}\scriptsize{$\pm$1.7} & 
        {42.7}\scriptsize{$\pm$0.8} & {40.4}\scriptsize{$\pm$1.8}  \\
        Subset (seen) & 
        89.9\scriptsize{$\pm$0.7} & 88.1\scriptsize{$\pm$0.5} & 
        90.0\scriptsize{$\pm$0.8} & 88.1\scriptsize{$\pm$0.4} & 
        56.2\scriptsize{$\pm$1.8} & 62.6\scriptsize{$\pm$1.5} & 
        49.6\scriptsize{$\pm$1.7} & 57.8\scriptsize{$\pm$1.8}  \\
        Subset (unseen) & 
        69.4\scriptsize{$\pm$0.4} & 66.4\scriptsize{$\pm$1.6} & 
        68.6\scriptsize{$\pm$0.5} & 66.0\scriptsize{$\pm$1.8} & 
        32.3\scriptsize{$\pm$0.7} & 33.1\scriptsize{$\pm$2.0} & 
        22.0\scriptsize{$\pm$1.1} & 24.3\scriptsize{$\pm$2.1}  \\
    \bottomrule
    \end{tabular}
    % }
    % \vspace{-0.1in}
\end{table*}

\subsection{Analysis on Seen or Unseen Rule Texts}
To investigate the model performance on the new, unseen rule texts, we conduct a detailed analysis on the ``seen'' and ``unseen'' split of validation and test set.
As mentioned in Section \ref{sec:dataset}, ``seen'' denotes the subset samples with the relevant rule text used in the training phrase while ``unseen'' denotes the subset samples with completely new rule texts.
This analysis aims to test the real scenario that some newly added rule texts may not have any associated user training data (user initial question, user scenario, and dialog history).
Results are put in Table \ref{tab:snippet-analysis}. We find that \modelnameshort\ performs extremely well on the ``seen'' subset with almost 10\% improvement than the overall evaluation set.
However, the challenge lies in the ``unseen'' rule texts -- our \modelnameshort\ performs almost 10\% worse than the overall performance which suggests this direction remains further investigations.

% \subsection{Impact of the Retrieval Performance}

\section{Conclusion}
In this paper, we propose a new open-retrieval setting of conversational machine reading (CMR) on natural language rules and create the first dataset \orsharc\ on this new setting.
Furthermore, we propose a new model \modelnameshort\ for open-retrieval CMR.
\modelnameshort\ explicitly locates conditions in multiple retrieved rule texts, conducts entailment-oriented reasoning to assess the satisfaction states of each individual condition according to the user-provided information, and makes the decision.
Results on \orsharc\ demonstrate that \modelnameshort\ outperforms all single-passage \sota\ models as well as our strong \mprbt\ baseline.
We further conduct in-depth analyses to provide new insights into this new setting and our model.

% \section*{Acknowledgments}
% The work described in this paper was partially
% supported by following projects from the Research
% Grants Council of the Hong Kong Special Administrative Region, China: CUHK 2300174 (Collaborative Research Fund, No. C5026-18GF); CUHK
% 14210717 (RGC General Research Fund).

\bibliographystyle{IEEEtran}
\bibliography{mybibfile,anthology}

\vfill

\end{document}